# Intelligent Search Heuristics for Cost Based Scheduling

Murphy Choy
Michelle Cheong

**Abstract**

Nurse scheduling is a difficult optimization problem with multiple constraints. There is extensive research in the literature solving the problem using meta-heuristics approaches. In this paper, we will investigate an intelligent search heuristics that handles cost based scheduling problem. The heuristics demonstrated superior performances compared to the original algorithms used to solve the problems described in Li et. Al. (2003) and Ozkarahan (1989) in terms of time needed to establish a feasible solution. Both problems can be formulated as a cost problem. The search heuristic consists of several phrases of search and input based on the cost of each assignment and how the assignment will interact with the cost of the resources.

**Introduction**

Nurse scheduling is a critical part of the daily operation in the hospital. Any robust systems that can automate the process of personnel scheduling has the potential to impact the operation bottom line as well as improving the morale of the nurses. Building such a system has resulted in many research topics in management science and operations research. Unlike other employee scheduling problems, nurse scheduling presents a unique challenge as the hospital operates continuously and cannot afford any lapse in the staffing of the facilities. This presents more variables and constraints in terms of the type and number of shifts that can be allocated. While it is arguably similar to scheduling in manufacturing plants, the nurse scheduling problem solution can lead to great benefits to the hospital. In contrast with the scheduling problem in other fields, the nursing scheduling problem needs to satisfy a range of staff requirements which can be legal in nature such as skill type as well as shift combinations. Nurse scheduling problems also present a longer planning horizon compared to other similar scheduling problems.

There is extensive literature that deals with nurse scheduling and the number of such articles have increased over the years [7]. The field has generated widespread interest in the research community as many problems exhibit unique constraints due to the operating environment. This resulted in a proliferation of approaches to solving the problem. Techniques such as mathematical programming [10], artificial intelligence methods [12], expert systems [9] and knowledge based systems [2] have been studied and applied to the problem. The more successful approaches to the problem are derived from meta-heuristic techniques which include genetic algorithms [8], simulated annealing [3], tabu search [4] and neighborhood searches [5]. There are attempts to fuse both heuristic and mathematical programming resulting in new techniques such as Lagrangian-based heuristic [1] as well as hyper-heuristics [6]. Most of the paper in the literature addressed simplified or modified versions of the original problem even though there is progress in developing techniques that deal with real scenarios. It is critical that we develop techniques which are able to address the complexities of the real world scenarios where it can make an impact.

Nurse scheduling can be formulated as a cost constrained allocation problem where the assignments has to consider the hard constraints and soft constraints. Hard constraints are usually due to legislations or material resource restrictions. Soft constraints are usually derived due to staffs' preferences. Most of the techniques will attempt to solve the problem for the hard constraints before optimizing the soft constraints to obtain the optimum solution. In the case of cost constrained allocation problem, both the hard and soft constraints are coded as cost in varying degree of magnitude. By having the constraints coded as cost in varying magnitude, the cost based optimization can occur in a hierarchical manner. Thus, the cost constrained nurse scheduling problem is a multi-objective hierarchical combinatorial problem. While it is not practical in a real scenario to find a solution that will satisfy all the constraints, it is possible to find a solution that fits all the hard constraints and most of the soft constraints. Through careful assignment of the cost values, the problem can solved using either mathematical programming or heuristic. However, the cost values assignment is not objective in nature and the value assign might vary. Any robust and efficient scheduling system must be able to address the problem by rapidly generating a viable schedule with minimum cost and time needed.

In this paper, we proposed a new intelligent search heuristics that is designed to specifically to address problems formulated in the form of cost constrained optimization problem. We will use this approach to solve two classic cases in the literature, Li et. Al. and Ozkarahan. The key issue for both problems is the huge number of constraints imposed which reflects the situation in real world. Due to the complexity of the problem, we will decompose the problem into several cost components in a hierarchical manner and deploy the intelligent search heuristic to solve each layer of the problem sequentially. The key

The paper is organized in the following way. Section 2 present the general model in a hierarchical cost constrained optimization form with the key common hard constraints and soft constraints. We will then describe in detail the Intelligent Search Heuristic in Section 3 and the implementation in Section 4. In section 5 and 6, we will present the formulation of the Li et. Al. [11] and Ozkarahan [13] problems and solve them using the Intelligent Search Heuristic comparing the time taken to reach a reasonable solution compared to the original solution. In section 7, we will present our conclusion.

**Nurse Scheduling Problem General Constraints**

For a nurse scheduling solution or roster to be viable, all the hard constraints has to be satisfied. There are several major hard constraints (will henceforth be known as A type constraints):
1. A nurse cannot be assigned more than a single shift within a 24 hours period.
    a. A nurse can only work a single shift for any day.
    b. A nurse may not work on a night shift and a morning shift consecutively.
2. A nurse with certain skill qualification can only be substituted with another nurse having the same skill qualification.
3. The minimum number of nurse will need to be assigned for each shift of a day. Under or over coverage are not allowed.
4. A nurse must be given x days of rest for a work period of y days.
5. A nurse can only work for z number of consecutive days.
6. A nurse can only work a maximum number of q days of a shift type.
7. A nurse can only work a maximum number of y night shifts before l rest days.

Constraint A1 ensures that no nurse is forced to work multiple shifts in a day which result in the nurse not receiving sufficient rest and is both undesirable and dangerous. Constraint A2 implies that certain duties which require specialized skills will need to have same skill substitution. It also implies that non-skill oriented duties can be replaced by someone with specialized skills. Most countries have regulations that stipulate the wards' minimum staffing requirements. Even if the number of patients is very low, there are still minimal requirements for staffing. Constraint A3 covers the minimum staffing requirements. The constraint A4 enforces sufficient number of rest days are allocated to the nursing staffs which is necessary for their individual well being. A4 alone is insufficient to enforce proper amount of rest as it only prescribes the number of days to rest and not the arrangement of rest. Constraint A5 is needed to limit the maximum consecutive number of working days to ensure rest for the nursing staffs. Constraint A6 is needed to limit the maximum number of shift types to fairness among the nursing staffs. Constraint A7 is needed to ensure sufficient rest days after consecutive night shifts.

Any other rules which are not considered as hard constraints are soft constraints. While it is ideal to satisfy all of them, they can be violated if the violation provides a feasible solution. Sometime, multiple violations are needed to obtain a feasible solution. Whenever soft constraints are broken, penalties have to be imposed and they are directly proportional to the importance of the constraint and the severity of the impact on the viability of the schedule.

Occasionally, the employees get to specify the constraints they wish to be imposed on their schedule. This may take on the form of personal preference such as four-in-a-row work schedule, or multiple rest days together. This presents a very fluid environment where the number and variety of constraints can be very large. Each condition imposed can vary the problem in terms of size and complexity exponentially. Thus any heuristic designed to solve the problem has to be robust and effective in a multitude of problems. There are major soft constraints commonly encountered in the literature and the list presented below does not attempt and is not exhaustive in nature.

- Maximum number of weekends that can be assigned in a given period where weekend period is defined by the employees.
- Maximum number of consecutive weekends that can be assigned.
- No night shifts prior to off shift weekend.
- No split weekends. Example, either both weekend days are assigned or not assigned.
- Identical shift types over a weekend. For example, both Saturday and Sunday are assigned or not assigned.
- Minimum number of days off after consecutive night shifts.
- Valid numbers of consecutive shift types.
- Shift type successions. Afternoon shift should not be followed by morning shift.
- Maximum total number of assignments for all days of the week.
- Avoid a secondary skill being used by a nurse. Sometimes a nurse may be able to cover a shift which requires a specific skill but they may be reluctant to do so as it is not their preferred duty. An example would be the nurse matron not wanting to stand in for a regular nurse.

The exact form and implementation of the constraints is dependent on the environment that it is found in. In the next section, we will formulate the constraints and objective function in the form of cost based formulation.

**Nurse Scheduling Problem Cost Based Formulation**

The general form of the cost based nurse scheduling problem takes the form of a minimization problem (in the case of nurse costs). Let us first declare the various sets and parameters that will be used in the formulation.

Sets of Values

| | |
|---|---|
| S | Shifts of the day (1 – AM, 2- PM, 3 - MN) |
| N | Nurse |
| D | Days of the planning horizon |
| L | Leave |
| M | Demand |
| O | Supply |
| P | Preference |
| C | Cost |
| E | No of duties required |
| $S_1$ | Specific set $S_1$ |
| . | . |
| . | . |
| . | . |
| $S_n$ | Specific set $S_n$ |

Parameters

| | |
|---|---|
| $X_{nsd\ldots}$ | Assignment for nurse n, shift s of day d (with $S_1, \ldots, S_n$) |
| $L_{nd\ldots}$ | Leave for nurse n for day d (with $S_1, \ldots, S_n$) |
| $P_{nsd\ldots}$ | Cost Reduction for nurse n, shift s of day d (with $S_1, \ldots, S_n$) |
| $C_{nsd\ldots}$ | Cost for nurse n, shift s of day d (with $S_1, \ldots, S_n$) |
| $M_{sd\ldots}$ | Demand for shift s of day d (with $S_1, \ldots, S_n$) |
| $O_{sd\ldots}$ | Supply for shift s of day d (with $S_1, \ldots, S_n$) |
| $E_{ns\ldots}$ | No of shifts for nurse n for shift s (with $S_1, \ldots, S_n$) |

Since this case uses penalized objective functions, the objective function can be declared in the following forms.

$$Minimize \quad Total\ Cost = \sum_{1}^{n} \sum_{\ldots}^{\ldots} \sum_{1}^{s} \sum_{\ldots}^{\ldots} \sum_{1}^{d} \sum_{\ldots}^{\ldots} \{X_{n\ldots sd} C_{n\ldots sd} - X_{n\ldots sd} P_{n\ldots sd}\}$$

In all the cases above, the decision variable $X_{n\ldots sd}$ is defined as below.

$$X_{n\ldots sd} = \begin{cases} 1 & Nurse\ n\ works\ on\ shift\ s\ on\ day\ d\ (with\ S1, \ldots, Sn) \\ 0 & Otherwise \end{cases}$$

The first key constraint is the shift constraint. Any nurse is allowed to work no more than a single shift per working day.

$$\sum_{1}^{s} X_{n\ldots sd} \leq 1 \quad \forall\ n, \ldots, d \quad (C1)$$

The number of rest days for a schedule of d days can be expressed as the complement of the maximum number of y working days for any period of d days.

$$\sum_{1}^{s}\sum_{1}^{d} X_{n...sd} \leq y \qquad \forall\, n, ... \qquad (C2)$$

Given any K consecutive work days within the d days period, there must be at least 1 rest day.

$$\sum_{1}^{s}\sum_{k}^{k+4} X_{n...sd} < k \qquad \forall\, n, ...\ where\ k \leq d - k \qquad (C3)$$

After i consecutive night shifts, there must be at least 1 sleep day and h rest day.

$$\sum_{j}^{j+i-1} X_{n...,3,d} + \sum_{j+i}^{j+h+i}\sum_{1}^{s} X_{n...sd} \leq j \qquad \forall\, n, ...\ where\ j \leq d - i - h \qquad (C4)$$

Where $s = 3\ represents\ night\ shift$

For nurse n who has applied for leave on day d,

$$\sum_{1}^{s} X_{n...sd} \leq L_{n...d} \qquad \forall\, n, d, ...\, S_1, ..., S_n \qquad (C5)$$

$$L_{n...d} = \begin{cases} 0 & Nurse\ n\ takes\ leave\ on\ day\ d\ (with\ S1, ..., Sn) \\ 1 & Otherwise \end{cases}$$

Where $S_1,..., S_n$ represent other peripheral requirements.

Consecutive shifts Night-Morning are not permitted in the schedule.

$$X_{n...1,d+1} + X_{n...3,d} \leq 1 \qquad \forall\, n, d, ...\, S_1, ..., S_n \qquad (C6)$$

Where $s = 3\ represents\ night\ shift$ and $s = 1\ represents\ morning\ shift$

The number of nurses in a ward must satisfy the following constraint,

$$\sum_{1}^{n} X_{n...sd} = O_{s...d} \qquad \forall\, n, s, ...\, S_1, ..., S_n \qquad (C7)$$

$$O_{s...d} \geq M_{s...d} \qquad \forall\, s, d, ...\, S_1, ..., S_n \qquad (C8)$$

Where $S_1,..., S_n$ represent other peripheral requirements.

Senior nurses with specialty training can operate as normal nurses.

$$\{O_{s...t+1,w+1,d} - M_{s...t+1,w+1,d}\} + O_{s...twd} \geq M_{s...twd} \qquad \forall\, s, d, t, w, ...\, S_1, ..., S_n \qquad (C9)$$

The last constraint is the number of shifts that a nurse needs to work. There are specified numbers of shifts that a nurse has to work and they need to be satisfied.

$$\sum_{1}^{d} X_{n...sd} = E_{ns...} \quad \forall\, n, ..., s \qquad (C10)$$

**Intelligent Search Heuristic**

The intelligent search heuristic is a simple and deterministic search algorithm that searches for imbalances in the constraints and cost to minimize the objective function. The algorithm comprises of several independent steps which mimics the human's ability to identify shortcomings in the schedules. The intelligent search heuristic follows a hierarchical order of search for optimal solution where the hierarchy is determined by the importance of the constraints.

The first step in the intelligent search heuristic is to order the search patterns in terms of the importance to the problem that takes the form of cost reduction. All the hard constraints will be assigned a high cost reduction in the same magnitude to ensure that they are all given the same priority in the hierarchy of constraints. Any violations of the hard constraints will earn a massive penalty to the objective function. This forces the algorithm to actively satisfy the hard constraints before the soft constraints. In certain formulation, the cost reduction can be summarized as total cost reduction for an assignment in the schedule.

Once the most important constraint has been found, the algorithm begins to search for the most imbalanced allocation and assigned a nurse for that position. The imbalance can be defined as the constraint allocation which has the highest penalty value among the allocations. An example will be the number of nurses assigned for a day. If there is a day which has more assignments needed than any other days, then there is an imbalance. In the case where there is no imbalance, the algorithm will select the first of the constraint allocation. To determine the most appropriate nurse to be assigned to the position, the algorithm will search for the combination of assignment and nurse which will result in the greatest drop in the cost. After the assignment, the algorithm will then check for over assignment problem where the nurse has been over assigned to too many duties or that shift of the day has been assigned too many staffs.

The algorithm will then be repeated for each layer of constraints in the hierarchy until there can be no new assignments or any assignments will result in an increase in the cost function. The intelligent search heuristic is described in pseudo-code form below.

1. Initialization
    a. Unassigned all the nurses.
    b. Input the cost for the various constraints.
2. Search for constraints
    a. Search for the constraint with the highest cost reduction.
    b. Search within the constraint allocation for the highest imbalance.
        i. Search for the highest number of assignment needed for the constraint.
        ii. If there is no imbalance, select the first allocation.

   c. Identify the position.
3. Assignment search
   a. For the position, search through the nurses' assignment for that position.
   b. Select the nurse with the highest cost value that has not been assigned to that position.
   c. Assign the nurse.
4. Condition checks
   a. Check that objective function has been reduced.
   b. Check that the nurse has not been over assigned.
     i. If nurse is over assigned, remove the previous assignment.
   c. Check that the shift has not been over assigned.
     i. If the shift is over assigned, search for the nurse with the minimum cost and unassigned the nurse from that shift.
   d. Check for open assignments.
     i. If there are open assignments, go to step 2-4.
     ii. If there are no open assignments, terminate search.

The algorithm will be implemented in the open source spreadsheet software, Libreoffice Calc to solve the benchmark example of Ozkarahan and Li et. Al.. The experiments were performed using a PC with Centrino Duo processor and Windows 7 operating system. In the next section, we will evaluate the algorithm in solving the Ozkarahan Case.

**Case Study 1: Ozkarahan, 1989**

In the original paper, the data is taken from Luke's Medical Center in Phoenix, Arizona. The objective of the scheduling problem is to minimize the weekly labor cost of the nurses inclusive of the nurse frustration cost due to understaffing or overstaffing. The first constraint of the above formulation requires that the number of shifts that a nurse has to serve has to be met. The second constraint requires that the number of nurses staffed for the period of each day to be equal to the demand and over and understaffing are penalized. The last constraint limits the shifts to the total available workers where the day has two independent 12-hour periods. There were some days where the nurses has requested days off. This is incorporated into the model.

The model is then implemented in Libreoffice Calc and shown below.

Figure 1. Screenshot of the implementation in LibreOffice Calc. Initial starting point of the Ozkarahan problem

The problem is then solved using the Intelligent Search Heuristics with the results shown below.

| Nurse | Cost | 1 A | 1 M | 2 A | 2 M | 3 A | 3 M | 4 A | 4 M | 5 A | 5 M | 6 A | 6 M | 7 A | 7 M | Final |
|---|---|---|---|---|---|---|---|---|---|---|---|---|---|---|---|---|
| AID12 | 600 | 0 | 0 | 0 | 0 | 0 | 0 | 0 | 0 | 0 | 0 | 1 | 0 | 1 | 0 | 0 |
| AID13 | 900 | 0 | 0 | 0 | 0 | 1 | 0 | 1 | 0 | 1 | 0 | 0 | 0 | 0 | 0 | 0 |
| AID14 | 600 | 1 | 0 | 1 | 0 | 0 | 0 | 0 | 0 | 0 | 0 | 0 | 0 | 0 | 0 | 0 |
| RN1 | 1500 | 0 | 0 | 0 | 0 | 0 | 1 | 0 | 1 | 0 | 1 | 0 | 1 | 0 | 1 | 0 |
| RN10 | 900 | 0 | 0 | 0 | 0 | 1 | 0 | 1 | 0 | 1 | 0 | 0 | 0 | 0 | 0 | 0 |
| RN11 | 900 | 1 | 0 | 1 | 0 | 0 | 0 | 0 | 0 | 0 | 0 | 0 | 1 | 0 | 0 | 0 |
| RN2 | 1800 | 0 | 1 | 0 | 1 | 0 | 0 | 0 | 1 | 0 | 1 | 0 | 1 | 0 | 1 | 100 |
| RN3 | 900 | 0 | 1 | 0 | 1 | 0 | 1 | 0 | 0 | 0 | 0 | 0 | 0 | 0 | 0 | 0 |
| RN4 | 900 | 0 | 0 | 0 | 0 | 0 | 0 | 1 | 0 | 1 | 0 | 1 | 0 | 0 | 0 | 0 |
| RN5 | 1200 | 1 | 0 | 0 | 0 | 1 | 0 | 0 | 0 | 0 | 0 | 1 | 0 | 1 | 0 | 0 |
| RN6 | 600 | 0 | 0 | 1 | 0 | 0 | 0 | 0 | 0 | 1 | 0 | 0 | 0 | 0 | 0 | 0 |
| RN7 | 900 | 0 | 0 | 1 | 0 | 1 | 0 | 1 | 0 | 0 | 0 | 0 | 0 | 0 | 0 | 0 |
| RN8 | 1200 | 1 | 0 | 0 | 0 | 0 | 0 | 0 | 0 | 1 | 0 | 1 | 0 | 1 | 0 | 0 |
| RN9 | 1200 | 1 | 0 | 1 | 0 | 1 | 0 | 1 | 0 | 0 | 0 | 0 | 0 | 0 | 0 | 0 |

Figure 2. Screenshot of the final solution of the Ozkarahan problem

The algorithm takes 1.4 seconds on the stopwatch to complete the search for a feasible solution. The time taken is then compared to the benchmark time taken for original solution.

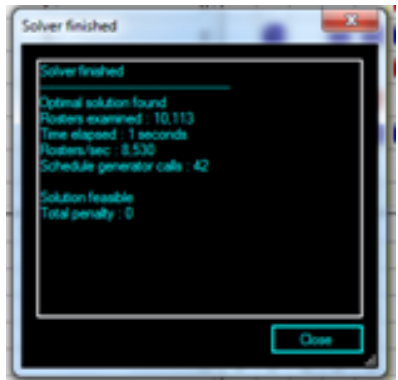

Figure 3. Screenshot of the final solution of the Ozkarahan problem using the software, roster booster

Using the benchmarking software, the diagnostics shows that the feasible solution is obtained after 1 second although the author recorded 1.7 seconds on the stopwatch. It is unclear whether the difference is due to the display of result. The algorithm is slightly faster than the benchmark. In the next section, we will review another benchmark case using the intelligent search algorithm.

**Case Study 2: Li et. Al., 2003**

The objective of the scheduling problem is to minimize the weekly labor cost of the nurses inclusive of the nurse frustration cost due to understaffing or overstaffing. It has similar constraints as the first two constraints above. However, for the third constraint, the problem has 3 shifts as opposed to 2 shifts. The model also incorporates more soft constraints such as leaves as well as shift requests.

The model is implemented in Libreoffice Calc and shown below.

| Nurse | Assignment | | | | | | | | | | | | | | | | | | | | | Cost |
|---|---|---|---|---|---|---|---|---|---|---|---|---|---|---|---|---|---|---|---|---|---|---|
| | 1 | | | 2 | | | 3 | | | 4 | | | 5 | | | 6 | | | 7 | | | |
| | A | P | M | A | P | M | A | P | M | A | P | M | A | P | M | A | P | M | A | P | M | |
| 1 | 0 | 0 | 0 | 0 | 0 | 0 | 0 | 0 | 0 | 0 | 0 | 0 | 0 | 0 | 0 | 0 | 0 | 0 | 0 | 0 | 0 | 50 |
| 2 | 0 | 0 | 0 | 0 | 0 | 0 | 0 | 0 | 0 | 0 | 0 | 0 | 0 | 0 | 0 | 0 | 0 | 0 | 0 | 0 | 0 | 40 |
| 3 | 0 | 0 | 0 | 0 | 0 | 0 | 0 | 0 | 0 | 0 | 0 | 0 | 0 | 0 | 0 | 0 | 0 | 0 | 0 | 0 | 0 | 50 |
| 4 | 0 | 0 | 0 | 0 | 0 | 0 | 0 | 0 | 0 | 0 | 0 | 0 | 0 | 0 | 0 | 0 | 0 | 0 | 0 | 0 | 0 | 50 |
| 5 | 0 | 0 | 0 | 0 | 0 | 0 | 0 | 0 | 0 | 0 | 0 | 0 | 0 | 0 | 0 | 0 | 0 | 0 | 0 | 0 | 0 | 30 |
| 6 | 0 | 0 | 0 | 0 | 0 | 0 | 0 | 0 | 0 | 0 | 0 | 0 | 0 | 0 | 0 | 0 | 0 | 0 | 0 | 0 | 0 | 60 |
| 7 | 0 | 0 | 0 | 0 | 0 | 0 | 0 | 0 | 0 | 0 | 0 | 0 | 0 | 0 | 0 | 0 | 0 | 0 | 0 | 0 | 0 | 50 |
| 8 | 0 | 0 | 0 | 0 | 0 | 0 | 0 | 0 | 0 | 0 | 0 | 0 | 0 | 0 | 0 | 0 | 0 | 0 | 0 | 0 | 0 | 50 |
| 9 | 0 | 0 | 0 | 0 | 0 | 0 | 0 | 0 | 0 | 0 | 0 | 0 | 0 | 0 | 0 | 0 | 0 | 0 | 0 | 0 | 0 | 30 |
| 10 | 0 | 0 | 0 | 0 | 0 | 0 | 0 | 0 | 0 | 0 | 0 | 0 | 0 | 0 | 0 | 0 | 0 | 0 | 0 | 0 | 0 | 40 |
| 11 | 0 | 0 | 0 | 0 | 0 | 0 | 0 | 0 | 0 | 0 | 0 | 0 | 0 | 0 | 0 | 0 | 0 | 0 | 0 | 0 | 0 | 40 |
| 12 | 0 | 0 | 0 | 0 | 0 | 0 | 0 | 0 | 0 | 0 | 0 | 0 | 0 | 0 | 0 | 0 | 0 | 0 | 0 | 0 | 0 | 30 |
| 13 | 0 | 0 | 0 | 0 | 0 | 0 | 0 | 0 | 0 | 0 | 0 | 0 | 0 | 0 | 0 | 0 | 0 | 0 | 0 | 0 | 0 | 40 |
| 14 | 0 | 0 | 0 | 0 | 0 | 0 | 0 | 0 | 0 | 0 | 0 | 0 | 0 | 0 | 0 | 0 | 0 | 0 | 0 | 0 | 0 | 30 |
| 15 | 0 | 0 | 0 | 0 | 0 | 0 | 0 | 0 | 0 | 0 | 0 | 0 | 0 | 0 | 0 | 0 | 0 | 0 | 0 | 0 | 0 | 40 |

Figure 4. Screenshot of the implementation in LibreOffice Calc. Initial starting point of the Li et. Al. problem

The problem is then solved using the Intelligent Search Heuristics with the results shown below.

| Nurse | Assignment | | | | | | | | | | | | | | | | | | | | | Cost |
|---|---|---|---|---|---|---|---|---|---|---|---|---|---|---|---|---|---|---|---|---|---|---|
| | 1 | | | 2 | | | 3 | | | 4 | | | 5 | | | 6 | | | 7 | | | |
| | A | P | M | A | P | M | A | P | M | A | P | M | A | P | M | A | P | M | A | P | M | |
| 1 | 0 | 0 | 1 | 0 | 1 | 0 | 0 | 0 | 0 | 1 | 0 | 0 | 0 | 0 | 1 | 0 | 0 | 0 | 1 | 0 | 0 | 50 |
| 2 | 0 | 0 | 1 | 0 | 1 | 0 | 1 | 0 | 0 | 0 | 0 | 0 | 0 | 0 | 1 | 0 | 0 | 0 | 1 | 0 | 0 | 40 |
| 3 | 1 | 0 | 0 | 0 | 0 | 0 | 0 | 0 | 0 | 1 | 0 | 0 | 0 | 1 | 0 | 1 | 0 | 0 | 0 | 0 | 0 | 50 |
| 4 | 1 | 0 | 0 | 1 | 0 | 0 | 0 | 0 | 0 | 1 | 0 | 0 | 0 | 1 | 0 | 0 | 0 | 0 | 1 | 0 | 0 | 50 |
| 5 | 0 | 1 | 0 | 1 | 0 | 0 | 0 | 0 | 1 | 0 | 0 | 0 | 1 | 0 | 0 | 0 | 0 | 0 | 0 | 0 | 0 | 30 |
| 6 | 0 | 0 | 0 | 0 | 0 | 1 | 0 | 0 | 0 | 1 | 0 | 0 | 1 | 0 | 0 | 0 | 1 | 0 | 0 | 0 | 0 | 60 |
| 7 | 1 | 0 | 0 | 0 | 0 | 0 | 1 | 0 | 0 | 0 | 0 | 0 | 0 | 1 | 0 | 0 | 0 | 0 | 0 | 0 | 1 | 50 |
| 8 | 0 | 1 | 0 | 0 | 1 | 0 | 0 | 0 | 0 | 0 | 0 | 0 | 0 | 1 | 0 | 0 | 0 | 1 | 0 | 1 | 0 | 50 |
| 9 | 0 | 0 | 1 | 0 | 1 | 0 | 0 | 0 | 0 | 0 | 0 | 0 | 0 | 1 | 0 | 1 | 0 | 0 | 0 | 0 | 0 | 30 |
| 10 | 0 | 0 | 0 | 0 | 1 | 0 | 1 | 0 | 1 | 0 | 0 | 0 | 0 | 0 | 0 | 0 | 0 | 0 | 0 | 0 | 0 | 40 |
| 11 | 1 | 0 | 0 | 0 | 1 | 0 | 1 | 0 | 0 | 0 | 1 | 0 | 0 | 0 | 0 | 0 | 0 | 0 | 0 | 0 | 0 | 40 |
| 12 | 0 | 0 | 0 | 0 | 0 | 0 | 0 | 0 | 1 | 0 | 0 | 0 | 1 | 0 | 0 | 0 | 1 | 0 | 1 | 0 | 0 | 30 |
| 13 | 0 | 1 | 0 | 1 | 0 | 0 | 0 | 1 | 0 | 0 | 0 | 0 | 1 | 0 | 0 | 0 | 0 | 0 | 0 | 1 | 0 | 40 |
| 14 | 0 | 0 | 0 | 0 | 0 | 0 | 0 | 0 | 0 | 0 | 0 | 0 | 0 | 0 | 0 | 0 | 0 | 0 | 0 | 0 | 0 | 30 |
| 15 | 0 | 1 | 0 | 0 | 0 | 0 | 1 | 0 | 0 | 0 | 1 | 0 | 0 | 0 | 0 | 0 | 0 | 0 | 0 | 0 | 1 | 40 |
| 16 | 0 | 1 | 0 | 0 | 0 | 0 | 0 | 1 | 0 | 0 | 0 | 1 | 0 | 0 | 0 | 1 | 0 | 1 | 0 | 0 | 0 | 60 |

Figure 5. Screenshot of the final solution of the Li et. Al. problem

The algorithm takes 4.4 seconds on the stopwatch to complete the search for a feasible solution. The time taken is then compared to the benchmark time taken for original solution.

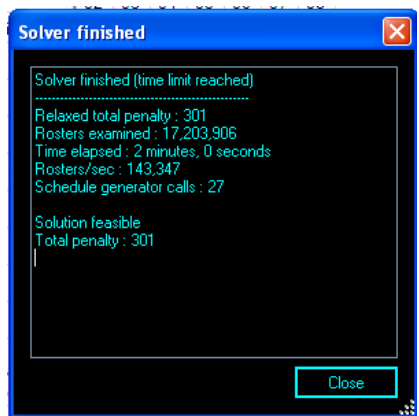

Figure 6. Screenshot of the final solution of the Li et. Al. problem using the software, roster booster

Using the benchmarking software, the diagnostics shows that the feasible solution is obtained after 2 minutes. The algorithm is much faster than the benchmark.

**Conclusion**

The Intelligent Search heuristic algorithm described has been shown to be a direct, simple but effective approach for cost base nurse scheduling problem. It is a viable and comparable alternative to the existing techniques used in Li et. Al. and Ozkarahan. The Intelligent Search heuristic algorithm has been shown to find a feasible schedule faster than the algorithms used to search for the previous problems.

Even though the results produced by this algorithm are fast and feasible, there are some areas in which it could be improved. The selection of the nurse to assign based on the cost alone might not be the most effective approach. It is possible that there is a more effective approach of selection takes into consideration more factors. Another major point of contention is the cleaning up of excess assignment. The selection criteria can be studied in greater in depth to evaluate the most best measure.
The current search mechanism is deterministic and elements of randomness injected into the system might yield a better solution.